\documentclass{article}

\PassOptionsToPackage{numbers}{natbib}

\usepackage[preprint]{neurips_2024}

\usepackage[utf8]{inputenc} 
\usepackage[T1]{fontenc}    
\usepackage{hyperref}       
\usepackage{url}            
\usepackage{booktabs}       
\usepackage{amsfonts}       
\usepackage{nicefrac}       
\usepackage{microtype}      
\usepackage{xcolor}         
\usepackage{amsmath}        
\usepackage{graphicx}       
\usepackage{subcaption}     
\usepackage{algorithm}
\usepackage{tabularx}       
\usepackage{algpseudocode}
\usepackage{float}
\title{Probabilistic Curriculum Learning for Goal-Based Reinforcement Learning}

\author{%
  Llewyn Salt\\
  Electrical Engineering and Computer Science\\
  University of Queensland\\
  Brisbane, Australia\\
  \texttt{llewyn.salt@gmail.com} \\
  \And
  Marcus Gallagher\\
  Electrical Engineering and Computer Science\\
  University of Queensland\\
  Brisbane, Australia\\
  \texttt{marcusg@eecs.uq.edu.au} \\
}

\begin{document}

\maketitle

\begin{abstract}
  Reinforcement learning (RL) --- algorithms that teach artificial agents to interact with environments by maximising reward signals --- has achieved significant success in recent years. These successes have been facilitated by advances in algorithms (e.g., deep Q-learning, deep deterministic policy gradients, proximal policy optimisation, trust region policy optimisation, and soft actor-critic) and specialised computational resources such as GPUs and TPUs. One promising research direction involves introducing goals to allow multimodal policies, commonly through hierarchical or curriculum reinforcement learning. These methods systematically decompose complex behaviours into simpler sub-tasks, analogous to how humans progressively learn skills (e.g. we learn to run before we walk, or we learn arithmetic before calculus). However, fully automating goal creation remains an open challenge. We present a novel probabilistic curriculum learning algorithm to suggest goals for reinforcement learning agents in continuous control and navigation tasks.
\end{abstract}

\section{Introduction}
Reinforcement learning (RL) --- algorithms that teach artificial agents to interact optimally with their environments by maximising reward signals~\cite{sutton2018reinforcement} --- has achieved remarkable success in recent years, notably in discrete and zero-sum games such as Atari~\cite{mnih2015human}, Go~\cite{silver2017mastering}, and Starcraft~\cite{vinyals2019grandmaster}. These achievements have been driven primarily by novel deep RL algorithms, including deep Q-learning~\cite{mnih2015human}, deep deterministic policy gradients (DDPG)\cite{lillicrap2015continuous}, proximal policy optimisation (PPO)\cite{schulman2017proximal}, trust region policy optimisation (TRPO)\cite{schulman2015trust}, and soft actor-critic (SAC)\cite{haarnoja2018soft}, combined with advances in computing hardware like GPUs and TPUs~\cite{jouppi2017datacenter}.

However, translating these successes to continuous environments remains challenging. In physical systems, unlike games that often have singular, explicit objectives (e.g., maximising score), RL must frequently address more diverse, nuanced goals—such as positioning a robot or controlling a lift precisely. Goal-based reinforcement learning aligns RL closely with optimal control~\cite{bertsekas2019reinforcement}, enabling agents to learn a variety of behaviours simultaneously, crucial for applications in autonomous vehicles, robotics, and flexible game AI~\cite{chane2021goal,tang2024deep, sikora2020overview,tang2024deep}. Yet, this broader goal space significantly complicates learning, especially in continuous environments, demanding more efficient strategies~\cite{klink2021probabilistic}.

Curriculum learning, which sequences tasks from simple to complex, offers one promising approach to efficiently mastering complex, multi-goal scenarios~\cite{bengio2009curriculum,narvekar2020curriculum, HeldGFA17, portelas2020automatic, gupta2022extending}. Still, existing approaches to automatically generating curricula face notable limitations: some require restrictive goal initialisations~\cite{florensa2018automatic}, while others rely on constrained Gaussian distributions to ensure training stability~\cite{klink2021probabilistic}. Further, evaluations often narrowly focus on singular target goals, contradicting the multi-goal paradigm's spirit.

To address these limitations, we propose a novel curriculum learning algorithm that explicitly models task difficulty using a probabilistic approach, enabling dynamic selection of goals that are neither trivial nor prohibitively challenging. By predicting an agent's likelihood of successfully achieving goals, our approach filters goals to a difficulty level suitable for efficient policy learning, whilst not requiring restrictive goal initialisation or constraining the probability distribution. We evaluate our algorithm in continuous control and navigation tasks, comparing performance to a baseline uniform curriculum. Our experiments specifically investigate whether probabilistically-driven goal selection leads to improved learning efficiency, generalisation across multiple goals, and improved performance in longer time horizon tasks.

\section{Background}
\subsection{Reinforcement Learning}
RL involves learning optimal sequences of actions in interactive environments modelled by Markov decision processes (MDPs)~\cite{sutton2018reinforcement}. At each step, an agent observes state $s_t$, performs action $a_t$, and receives reward $r_t$, with transitions governed by a probability function $\mathcal{T}(s_t,a_t,s_{t+1})$. RL's objective is to learn a policy $\pi$ that maximises cumulative discounted reward $R_t = \sum_{i=t}^{T}\gamma^{i-t}r_i$. Optimal policies are characterised by their action-value functions $Q^{\pi}(s_t,a_t) = \mathbb{E}[R_t | s_t,a_t]$, satisfying the Bellman optimality equation:
\begin{equation}
  \
   Q^\pi(s_t,a_t) = \mathbb{E}_{s_{t+1}}[r_{t+1} + \gamma Q^\pi(s_{t+1},\pi(s_{t+1}))]
\end{equation}

\subsection{Goal-Based Reinforcement Learning}
Goal-based RL generalises traditional RL by conditioning policies $\pi(s_t,g_t)$ and value functions $Q(s_t,a_t,g_t)$ explicitly on goals $g$, enabling agents to achieve diverse outcomes without extensive reward shaping~\cite{pmlr-v37-schaul15,NIPS2017_7090}. Goals are sampled as subsets of the state space, guiding policies across diverse tasks and facilitating learning in sparse-reward environments. Universal value function approximators (UVFAs) extend the $Q$-function to explicitly incorporate goals, thereby generalising the action-value function to multiple goal-conditioned reward functions $r_g(s_t,a_t,s_{t+1})$. This approach allows agents to flexibly adapt their policy according to the desired goals, significantly improving generalisation capabilities.

\subsection{Curriculum Learning}
Curriculum learning sequences tasks in increasing complexity, mirroring the structured progression humans naturally use when acquiring new skills~\cite{bengio2009curriculum,wang2021survey,narvekar2018learning,yin2024autonomous}. Although handcrafted curricula have shown efficacy in reinforcement learning (RL), manual curriculum design is typically tedious, subjective, and limited to specific scenarios~\cite{karpathy2012curriculum}. Consequently, recent efforts have focused on automated curriculum generation methods~\cite{forestier2017intrinsically, graves2017automated, svetlik2017automatic}.

Goal GAN~\cite{pmlr-v80-florensa18a} exemplifies one such automated approach, generating intermediate-difficulty goals, yet it requires careful goal initialisation and suffers from GAN-related training instabilities~\cite{creswell2018generative, barnett2018convergence,goodfellow2020generative}. Similarly, Self-Paced Learning (SPL) methods, such as Klink et al.'s probabilistic approach~\cite{klink2021probabilistic}, adaptively adjust task complexity but rely on narrow Gaussian context distributions to maintain stability, limiting flexibility and generalisation across broader goal spaces. Additionally, these methods typically evaluate performance against a single fixed goal, whereas in true multi-goal RL, the agent sequentially adapts to different goals—each episode or upon achieving the current target—further complicating training.

To address these limitations, we propose a novel probabilistic curriculum learning approach leveraging stochastic variational inference (SVI) to dynamically estimate task difficulty, facilitating stable and flexible goal selection suitable for scalable multi-goal reinforcement learning.

\subsection{Stochastic Variational Inference}
Stochastic Variational Inference (SVI) has emerged as a powerful technique for approximate Bayesian inference, particularly in scenarios where exact inference becomes computationally prohibitive due to model complexity or dataset size~\cite{hoffman2013stochastic, blei2017variational}. By transforming Bayesian inference into an optimisation problem, SVI approximates the posterior distribution using a simpler, variational distribution. Unlike traditional variational inference, SVI leverages stochastic optimisation by iteratively updating parameters based on gradients computed from random subsets (mini-batches) of data, significantly enhancing scalability and computational efficiency~\cite{kingma2013auto,lalchand2022generalised}. This efficiency has led to broad applicability across diverse fields, including probabilistic machine learning, deep generative models such as variational autoencoders, and large-scale natural language processing tasks~\cite{hoffman2015structured, zhang2018advances}.

\section{Probabilistic Curriculum Learning}
We present probabilistic curriculum learning (PCL) as a novel method to suggest goals for reinforcement learning agents in continuous control and navigation tasks.
\subsection{Formalisation of the Problem}
First we must clearly define some of the mathematical assumptions and definitions that we will use to formalise the algorithm.
\subsubsection{Linking Goals and States}
\label{ssec:curriculum:goallinkage}
We assume that there exists some function $f: \mathcal{S} \to \mathcal{G}$ such that we can then calculate the reward of reaching the desired goal as:
\begin{equation}
   r_g(s_{t+1},g_t) = \begin{cases}
                     1, \text{if } D(f(s_{t_1}),g_t)<\epsilon\\
                     0, \text{otherwise} 
                  \end{cases}
\end{equation}
where $D$ is some distance function like the Euclidean distance and $\epsilon << 1$ is some error margin as the probability of $f(s_t)$ being exactly equal to $g_t$ in continuous space is zero. In our experiments $\mathcal{G} \in \mathbb{R}^N$ is a subset of  $\mathcal{S}\in  \mathbb{R}^M$ where $N\leq M$ so  $f(s_t) = s_t\times \mathbf{I}^{M\times N}_R$  where $\mathbf{I_R}$ is a row reduced identity matrix and $s_t$ is an $M\times 1$ vector. If $\mathcal{S}=\mathcal{G}$ the $f(s_t) = \mathbb{I}\times = s_t$, but if $g_t$ only specifies some of the properties of $s$, e.g. suppose that robots state is specified by ${x,y,z}$, but we desire that the robot should be able to reach any arbitrary ${x,z}$ irrespective of $y$ then $f((x,y,z))=\begin{bmatrix}
   1& 0& 0\\
   0& 0& 1
\end{bmatrix}  \begin{bmatrix}
   x\\
   y\\
   z
\end{bmatrix} = \begin{bmatrix}
   x\\
   z
\end{bmatrix}$  this is the simplest way to model goals. Goals could potentially be a combination of states e.g a specific policy or something that is as yet unthought of, but is outside the scope of this paper.

\subsubsection{Probability Density Estimation for Goal Selection}
We propose a novel technique whereby the probability that a goal is successful ($g^s_t$) given the current policy, $\pi$: $P(g^s_t|\pi)$ is estimated. Stochastic variational inference techniques can be used to learn the probability density function $p(g^s_t|\pi)$. 

$p(g^s_t|\pi)$ can be estimated using any probability density estimator. However, as the policy $\pi$ is typically approximated using a function approximator, such as a neural network, we choose to characterise it through the state action pairs it experiences, $(s_t, a_t)$, where $a\sim \pi(s_t)$. 

We can estimate the pdf using any probability density estimator. We know that 
$g^s_t \in \mathbb{G}$ and $\mathbb{G} \subset \mathbb{S}$, so we can say that a successful goal given a state action pair is characterised by any 
subsequent state, $s_{t+N}$, within the same epoch. Therefore, we can assert that  
\begin{equation}
    \label{eq:goal:equiv_pdf}
    \begin{split}
        p(g^s_t|\pi) & \approx p(g^s_t|s_t,a_t) \\
                     & \approx p(s_{t+N}|s_t,a_t)
    \end{split}
\end{equation}
We can now sample candidate goals from the above distribution, $g^c_{t} \sim p(s_{t+1}|s_t,a_t)$. However, as we have some probability of sampling goals that
have a low probability of success we want to evaluate $P(g^c_{t}|s_t,a_t)$, as these are continuous random variables there is a zero probability that 
$g^c_{t}$ is exactly equal to a successful goal. We could assume some volume and calculate the goal probability as per Appendix~\ref{app:goal:probability}. For example if the model as a Gaussian mixture model then we can use Equation~\ref{eq:mvnprob}:
\begin{equation}
    \label{eq:mvnprob}
      P(g^c_{t}\in \mathbf{C}|s_t,a_t) = \sum_{j=1}^K\phi_j\frac{1}{\sqrt{(2\pi)^N|\Sigma_j|}}\prod_{i=1}^N\frac{\sqrt{\pi}\sqrt{\Sigma_{j_{ii}}}}{\sqrt{2}}(erf(\frac{g_{c_i}+\epsilon-\mu_{j_i}}
                                 {\sqrt{2}\sqrt{\Sigma_{j_{ii}}}})-erf(\frac{g_{c_i}-\epsilon-\mu_{j_i}}{\sqrt{2}\sqrt{\Sigma_{j_{ii}}}}))
\end{equation}
 
Where the volume is a hypercube $\mathbf{C}$ centred about $g_c$ with each dimension offset on either side of $g_c$ by $\epsilon>0$, and we also assume 
all $\Sigma_j$ are diagonal, and the goal dimensions are independent, see Appendix~\ref{app:goal:gaussianproof} for proof. The downside of the above is that the computation is limited to the Gaussian distributions with a specific volume to integrate over, and depending on the dimensionality of $\mathbf{C}$ it can be computationally expensive. Instead, we can look to using computational methods to estimate the probabilities to broaden the set of usable distributions. 

For a generic probability distribution we could utilise Monte Carlo Integration to estimate the probability of a goal being successful~\cite{geweke1996monte, kong2003theory}. However, this is computationally expensive. 

Instead, we decide to bound our pdf values by some quantile's upper and lower, we can then select from within this range either randomly or using some heuristic like min or max depending on how we set our bounds.

\begin{figure}[htbp]
    \begin{minipage}[t]{0.48\textwidth}
        \centering
        \includegraphics[height=3.9cm]{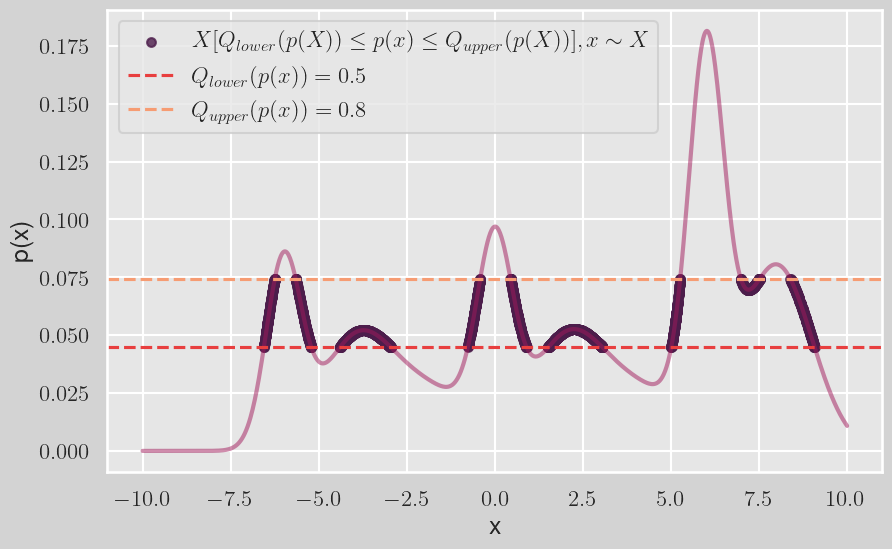}
        \caption{Illustration of the interaction between $Q_{upper}$ and $Q_{lower}$, the pdf, and goal selection.}
        \label{goal:shap:pdfillustration}
    \end{minipage}
    \hfill
    \begin{minipage}[t]{0.48\textwidth}
        \centering
        \includegraphics[height=3.9cm]{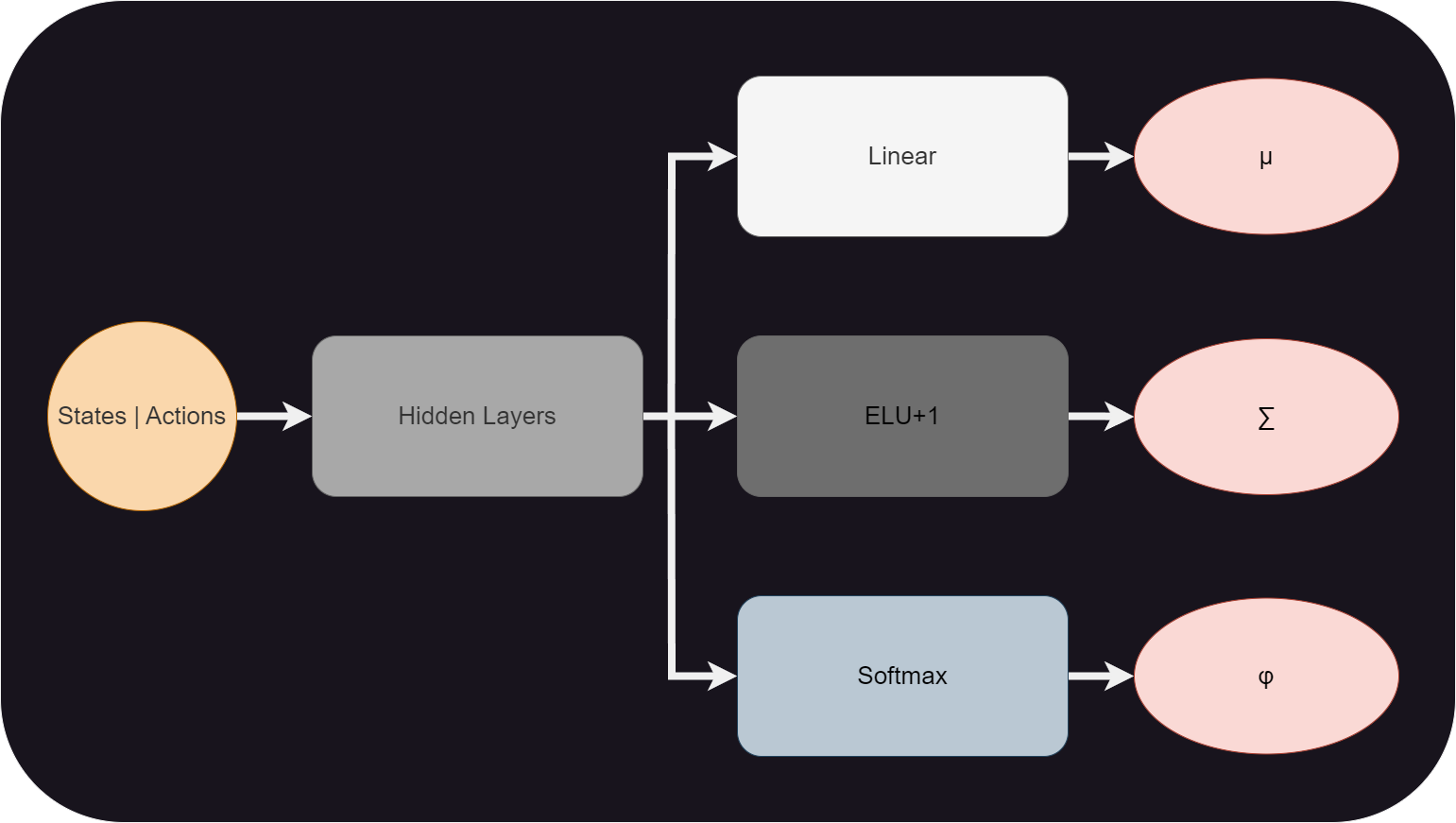}
        \caption{The deep mixture density network architecture.}
        \label{fig:goalsmdn}
    \end{minipage}
\end{figure}

\subsubsection{Mixture Density Network}
\label{sssect:mdn}
We use a mixture density network (MDN) to learn $\phi_j$, $\mu_j$, and $\Sigma_j$ and generate a Gaussian mixture model~\cite{bishop1994mixture,makansi2019overcoming}.  Figure~\ref{fig:goalsmdn} shows the network architecture of the Deep MDN used in our experiments, the hidden layer block uses a combination of non-linearity, batch-normalisation~\cite{bjorck2018understanding}, and dropout~\cite{baldi2013understanding}.

The only fixed part of this topology is the final layer which constrains the outputs to the ranges required for a Gaussian mixture model. The network could be changed to output the parameters of any distribution. The output shape of: $\phi_j$ is the number of mixtures $K$, and $\mu_j$ and $\Sigma_j$ is $K$ times the number of variables characterised by the goal. Stochastic Gradient Descent(SGD) is then used to minimise the negative log-likelihood loss given the mixture parameters:
\begin{equation}
   \label{eq:goal:mdnloss}
   L(s_{t+1}|s_t,a_t) = -\frac{1}{N}\sum^N_{i=1}\log(p(g_t|s_t,a_t))
\end{equation}
where $N$ is the minibatch size and $s_t,a_t,s_{t+1}$ are sampled from the agents experience. $s_{t+1}$ is converted into the goal $g_t$ as described in Section~\ref{ssec:curriculum:goallinkage}.

To prevent overfitting and modal collapse, we can also introduce some additional terms to the loss function. We have modified the loss function to be the sum of ELBO, L2-regularisation, and KL divergence which are useful in preventing modal collapse and overfitting:
\begin{equation}
    \label{eq:goal:mdnlosscomplete}
    L_\Theta(s_{t+1}|s_t,a_t) = -\frac{\lambda_1}{N}\sum^N_{i=1}\log(p(g|s_t,a_t)) + \lambda_2||\Theta||^2 + \lambda_3D_{KL}(q(s_{t+1})||p_\Theta(g|s_t,a_t))
\end{equation}
where $\Theta$ is represents the parameters of the deep MDN network, and $\lambda_1$, $\lambda_2$, and $\lambda_3$ are hyperparameters that control the strength of the ELBO loss, regularisation, and KL divergence terms. The regularisation term is useful in preventing overfitting and the KL divergence term is useful in preventing modal collapse.

\subsection{Algorithm}
\label{ssec:goal:algorithm}
Our goal sampling technique is straight forward, at the beginning of an episode we use the sample a set of goals $G$ from some distribution $\mathcal{D}$, which could be the probabilistic model, a uniform distribution, or something else. We then select the goal randomly from the set given by the probability density values that sit between a lower and upper quantile, i.e. $Q_{lower}(p(x|y))\leq p(x|y) \leq Q_{upper}(p(x|y))$ where $x$ is randomly sampled from the probabilistic model. After each step in the environment we store $s_t,a_t,s_{t+1}, g_t$  in a replay buffer. At each step we sample from the replay buffer and perform gradient descent on our probabilistic model to minimise the loss. At the end of each episode we select a new goal and repeat. Our algorithm is formally laid out in Algorithm~\ref{alg:goal:autogoalMDN}.

\begin{algorithm}[tb]
   \caption{Goal Generation using a probabilistic model for Reinforcement Learning}
   \label{alg:goal:autogoalMDN}
\begin{algorithmic}[1]
   \State {\bfseries Input:} an agent $\pi$, an environment $E$, a sampling distribution $\mathcal{D}$, a selection strategy $f(.)$, and a reward function $r_g$
   \State Randomly initialise probabilistic model, $\Theta$ 
   \State Initialise replay buffer $\mathcal{R}$
   \For{$step=1$ {\bfseries to} $max\_steps$}
      \State Sample initial state $s_0$ from $E$
      \If{epoch < warmup}
         \State Randomly select $g$ from $\mathcal{U}(s_{min},s_{max})$
      \Else
         \State Sample $G \in \mathcal{R}^{M,N}$ from $\mathcal{D}$
         \State Convert $s_0$ to $S_0 \in \mathcal{R}^{M,O}$ by repeating $s_0$ $M$ times
         \State Obtain $a \sim \pi(S_0, G)$
         \State Calculate $p_\Theta(G|s_t,a_t)$
         \State Pick $g \sim f(G[Q_{lower}<G<Q_{upper}])$ 
      \EndIf
      \While{terminal state not reached}
         \State Select action $a_t=\pi(s_t, g_t)$ 
         \State Execute action $a_t$ and observe a new state $s_{t+1}$
         \State Calculate the reward $r_t=r_g(s_{t+1},g)$ 
         \State Store experience ($s_t$, $a_t$, $r_t$, $s_{t+1}$, $g$)
         \State Update agent $\pi$ using its update rule
         \State Update probabilistic model by sampling a minibatch from $\mathcal{R}$ using eq.~\ref{eq:goal:mdnlosscomplete} and SGD
      \EndWhile
   \EndFor
\end{algorithmic}
\end{algorithm}

$f(.)$ is the selection strategy, which can be any method by which we select the goals once they have been filtered by the quantiles. Selection strategies and automatic quantile adjustment are discussed below.

\subsubsection{Goal Selection Strategies}
\label{sssect:goal:selection}
We develop the following candidates for the selection strategy: \textbf{uniform} We can select the goal randomly utilising a uniform distribution, which is the simplest and is the selection strategy used in experiments unless otherwise stated i.e. $f(.) \sim \text{Uniform}(1, N) $. 

\textbf{Weighted} selection using the normalised pdf values so they sum to 1 which informs a weighted random selection. Given a set of goals $\mathbf{g} = \{g_1, g_2, \ldots, g_N\}$ and their corresponding probabilities density values $\mathbf{p} = \{p_1, p_2, \ldots, p_N\}$, the probabilities are normalised to ensure they sum to 1, $p_i = \frac{p_i}{\sum_{j=1}^N p_j}$. If the normalisation sum is numerically close to zero, indicating degenerate probabilities, then the selected goal is chosen uniformly at random from the set of goals as per the \textbf{uniform} strategy. Thus, the goal-selection process is formally defined as:
\begin{equation}
    \label{eq:goal:selection:weighted}
    g^* = 
    \begin{cases}
    g_i, & \text{with probability } p_i = \frac{p_{i}}{\sum_{j=1}^N p_{j}}, \quad\text{if}\quad \sum_{j=1}^N p_{j} \geq 10^{-12} \\[10pt]
    g_i, & i \sim \text{Uniform}(1, N), \quad\text{otherwise}
    \end{cases}
\end{equation}

\textbf{multiweighted} selection strategy that considers multiple criteria: Given a set of goals $\mathbf{g}$ and their corresponding probabilities density values $\mathbf{p}$ as above, the combined score $p_i$ for each goal $g_i$ is computed as follows:
\begin{equation}
    S_i = \beta_1 \cdot U(g_i) + \beta_2 \cdot LP(g_i) + \beta_3 \cdot N(g_i)
\end{equation}

Where $U(g_i)$ is the Uncertainty of goal $g_i$, $LP(g_i)$ is the Learning Progress of goal $g_i$, and $N(g_i)$ is the Novelty of goal $g_i$. The terms are defined as follows:
\begin{align}
    U(g_i) &= p_i \cdot (1 - p_i) \\
    LP(g_i) &= |p_i - \text{old\_prob}(g_i)| \\
    N(g_i) &= \min_{g_j \in \text{known\_goals}} \| g_i - g_j \| 
\end{align}  
$\beta_1, \beta_2, \beta_3$ are the weights for Uncertainty, Learning Progress, and Novelty, respectively. The probabilities for sampling each goal are then normalised to sum to 1 as per the \textbf{weighted} strategy and the goal is sampled based on the distribution utilising the same Equation~\ref{eq:goal:selection:weighted}

\subsubsection{Adaptive Quantiles}
\label{sssect:goal:selection:adaptive}
The quantiles, $Q_{lower}$ and $Q_{upper}$, are adapted according to some success criteria. We collect a short-term memory of the past N goals and calculate the success rate, $SR$, the streak of successes, $s$, and a correction factor, $cf$. The short term memory, $\mathbf{a} = \{f(g_1), f(g_2), \ldots, f(g_N)\}$, contains a collection of binary values where:
\begin{align}
    f(g) &= \begin{cases}
                1 & \text{if goal is reached} \\
                0 & \text{otherwise}
            \end{cases}\\
    sr &= \frac{\sum_{i=1}^{N} a_i }{N} \text{ where } a_i \in \mathbf{a}\\
    s &= \max\{ k \mid f(g_{N - k + 1}) = f(g_{N - k + 2}) = \dots = f(g_N),\, 1 \leq k \leq N \}\\
    cf &= (1 - |sr - sr_{target}|) \times \alpha^s
\end{align}
Goal is reached is usually defined as the agent being within some tolerance of the goal which is typically defined by the environment. $\alpha$ is the factor by which we want the streak to exponentially impact the quantiles and $sr_{target}$ is the target success rate. The quantiles are then updated as follows:
\begin{equation}
    Q_{x} = \begin{cases}
                \min(min_{Q_x}, Q_x - \lambda \times cf) & \text{if } a_N = 1\\
                \max(max_{Q_x}, Q_x + \lambda \times cf) & \text{if } a_N = 0
    \end{cases}
\end{equation}
where $min_{Q_x}$ and $max_{Q_x}$ are the minimum and maximum values of the quantile, the $x$ from $Q_x$ is either upper or lower, $\lambda$ is the learning rate, and $a_N$ is the last element of the memory.

\subsection{Methodology}
\label{ssec:goal:methodology}
All experiments were conducted using the AlgOS framework~\cite{Salt2025}. SAC from Stable Baselines 3~\cite{stable-baselines3} is used as the agent to test the efficacy of the PCL in a DC Motor control environment, and a point robot navigation task~\cite{gymnasium_robotics2023github}. Both tasks are continuous control, with the DC Motor offering a single-input-single-output (SISO) control problem with no obstacles and the point robot navigation task offering a multi-input-multi-output (MIMO) control problem with obstacles.

AlgOS provides an optimisation interface via Optuna~\cite{optuna_2019}, which allows us to tune the numerous hyperparameters of both PCL and SAC using a tree parzen estimator~\cite{watanabe2023tree}, a form of Bayesian optimisation requiring fewer samples~\cite{salt2019parameter}. The bounds of the optimisation can be found in Table~\ref{tab:SACWeightedMDNExperiment} in Appendix~\ref{app:hyperparams}, which many of SAC's values were determined from Raffin et al.'s paper\cite{raffin2021smooth}. Additionally, due to compute resourcing constraints, we optimise over 150000 steps for all environments which is a relatively small number when compared to the number of steps used in the literature. However, we are not attempting to achieve maximum performance but rather explore where and how Algorithm~\ref{alg:goal:autogoalMDN} improves or diminishes the agent's capability to learn.

We will compare using an MDN probabilistic sampler trained to model $p(s_{t+N}|s_t,a_t)$ and the uniform sampler samples goals from the goal space as:
\begin{enumerate} 
    \item DC Motor: $x \sim U(a, b)$, where a and b are the minimum and maximum angular velocities of the motor.
    \item Point Maze: $x \sim \sum_{i=1}^{N} \frac{1}{N} \cdot U(\mathbf{l}_i, \mathbf{h}_i)$ where $N$ is the number of goals, $\mathbf{l}_i$ and $\mathbf{h}_i)$ are the upper and lower bounds of each cell in which the goal resides (2D vectors). 
\end{enumerate}

The goal is reached when: DC Motor: the agent stays within a tolerance 0.001 of the goal for 10 steps; Point Maze: the agent gets within 0.45 of the goal.

The algorithm's performance is measured using coverage, which is the percentage of the goals the agent can reach when evaluated. At evaluation the agent is tasked to reach a set of goals, $G = {g_1,...,g_N}$, four times. This ensures the agent can reach the same goal multiple times. Coverage is used as the objective metric for the hyperparameter optimisation. 

We will present the best three runs from each set of hyperparameter optimisations. We will also explore the effect of the adaptive quantile and sampling strategies. Unless stated otherwise, the \textbf{uniform} strategy with static quantiles is used.  

\section{Results and Discussion}
\label{sec:goal:results}
\subsection{DC Motor}
\begin{figure}[htbp]
    \centering
    \begin{subfigure}[t]{0.48\textwidth}
        \centering
        \includegraphics[height=4.3cm]{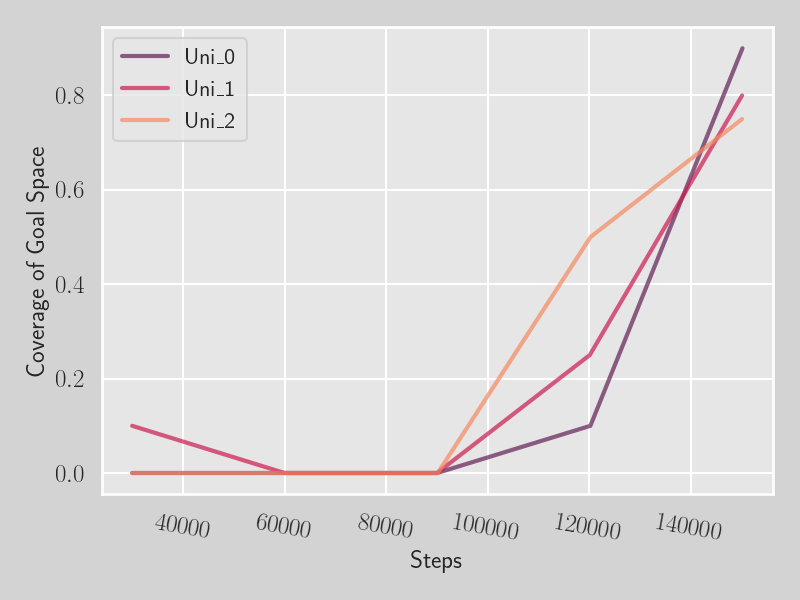}
        \caption{Uniform Coverage}
        \label{fig:goal:uni:coverage}
    \end{subfigure}
    \hfill
    \begin{subfigure}[t]{0.48\textwidth}
        \centering
        \includegraphics[height=4.3cm]{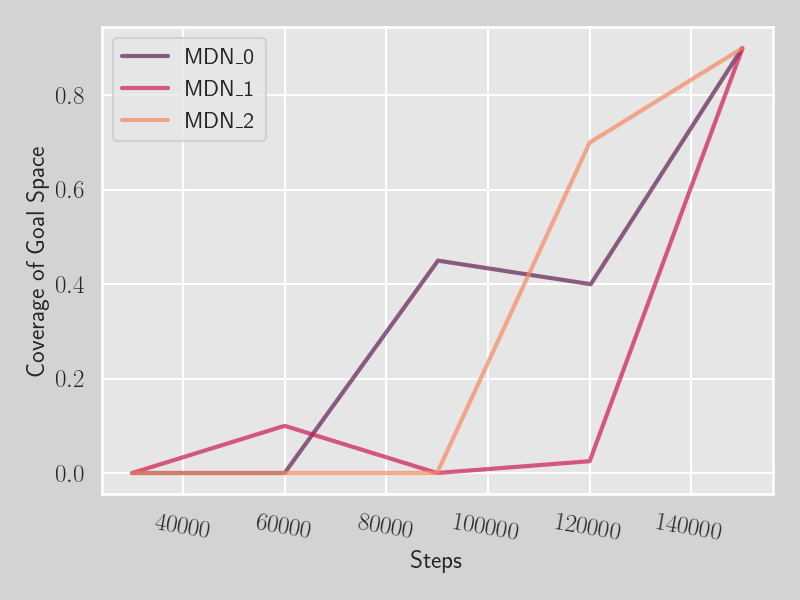}
        \caption{PCL Coverage}
        \label{fig:goal:mdn:coverage}
    \end{subfigure}
    \begin{subfigure}[t]{0.48\textwidth}
        \centering
        \includegraphics[height=3.9cm]{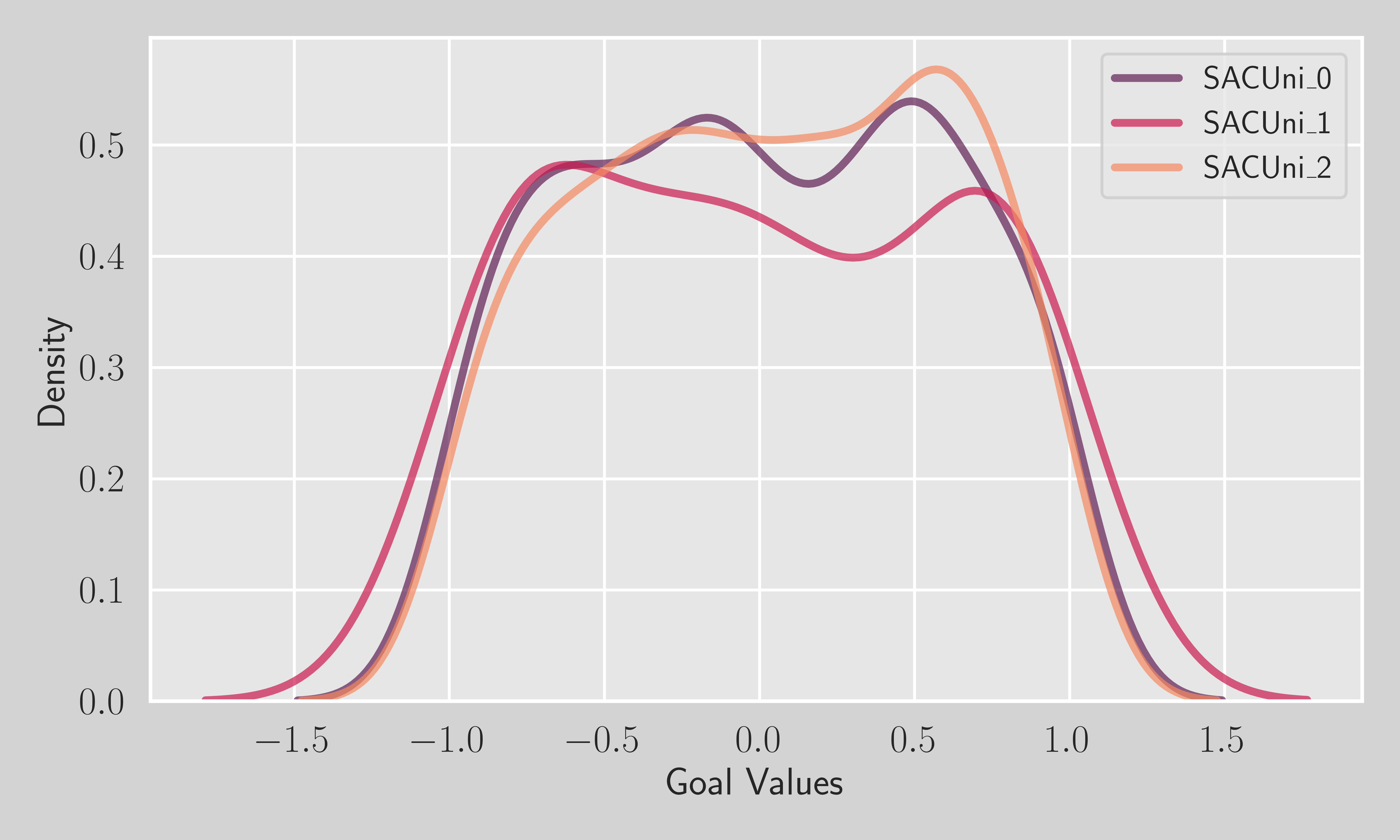}
        \caption{Uniform Distribution of Goals}
        \label{fig:goal:uni:dist}
    \end{subfigure}
    \hfill
    \begin{subfigure}[t]{0.48\textwidth}
        \centering
        \includegraphics[height=3.9cm]{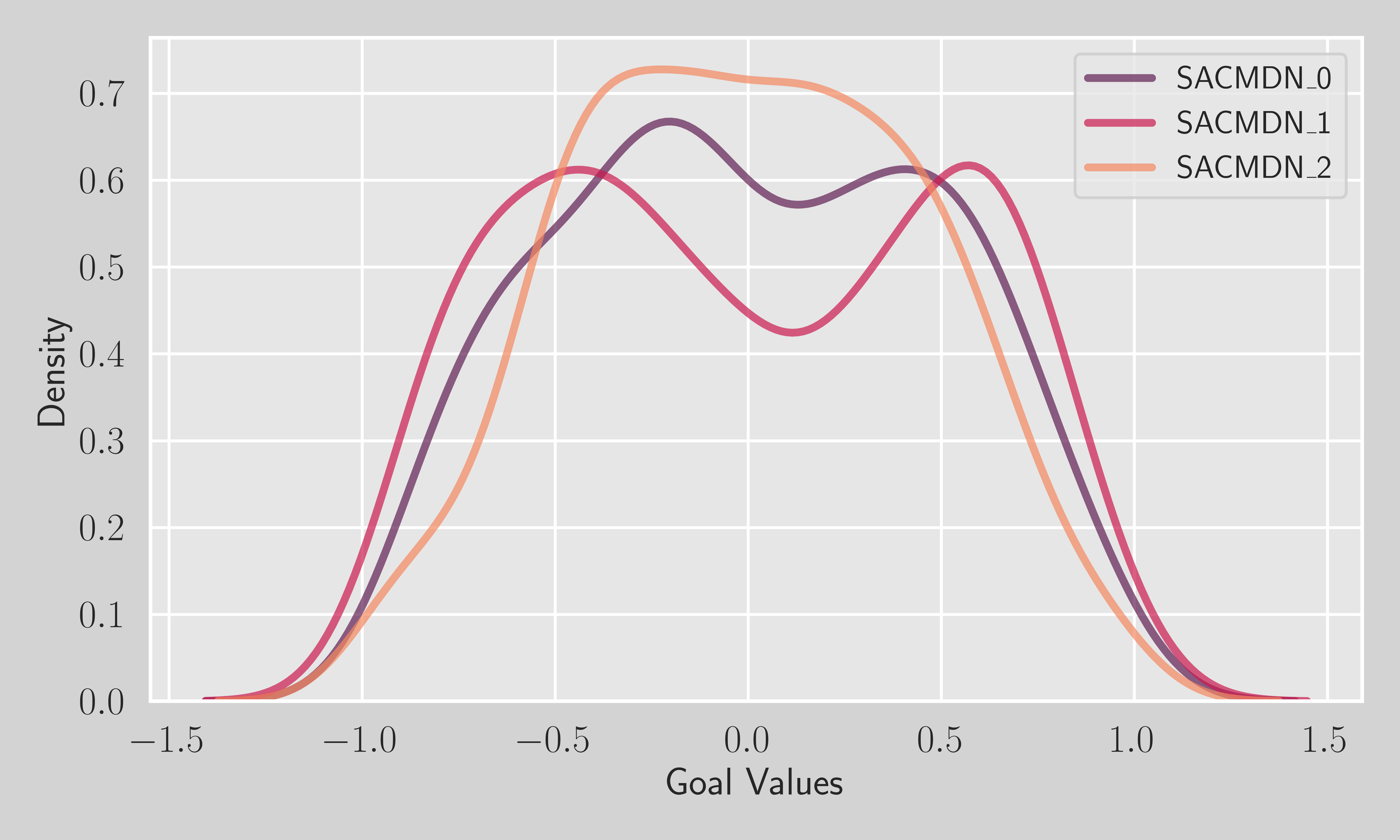}
        \caption{PCL Distribution of Goals}
        \label{fig:goal:mdn:dist}
    \end{subfigure}
    \caption{DC Motor Coverage and Distribution of Goals.}
    \label{fig:goal:dc_motor}
\end{figure}

Figure~\ref{fig:goal:dc_motor} shows the results of the DC Motor experiments, see Tables~\ref{tab:SACMDNDCMOTOR270125} and~\ref{tab:SACUNIDCMOTOR270125} for hyperparameters. The coverage is shown in the top row and the distribution of goals is shown in the bottom row. Both the uniform and PCL methods are able to achieve a coverage of 0.9, but the PCL is able to on all three runs. The distribution of goals for both the uniform and PCL methods is similar. It was assumed that for a simple problem like the DC motor the optimal curriculum would be a bimodal Gaussian distribution centred about $\pm0.5$ as close to 0 would be easy and $\pm1$ would be hard. Figures~\ref{fig:goal:uni:dist} and~\ref{fig:goal:mdn:dist} show that the best performers have similar distributions.

Figures~\ref{fig:goal:uni:coverage} and~\ref{fig:goal:mdn:coverage} show that the PCL method has non-zero coverage prior to 90,000 steps whereas the uniform method only increases after 90,000 steps, indicating that the PCL method increases training efficiency. 

\subsection{Point Maze}
\begin{figure}[htbp]
    \centering
    \begin{subfigure}[t]{0.245\textwidth}
        \includegraphics[height=3.9cm]{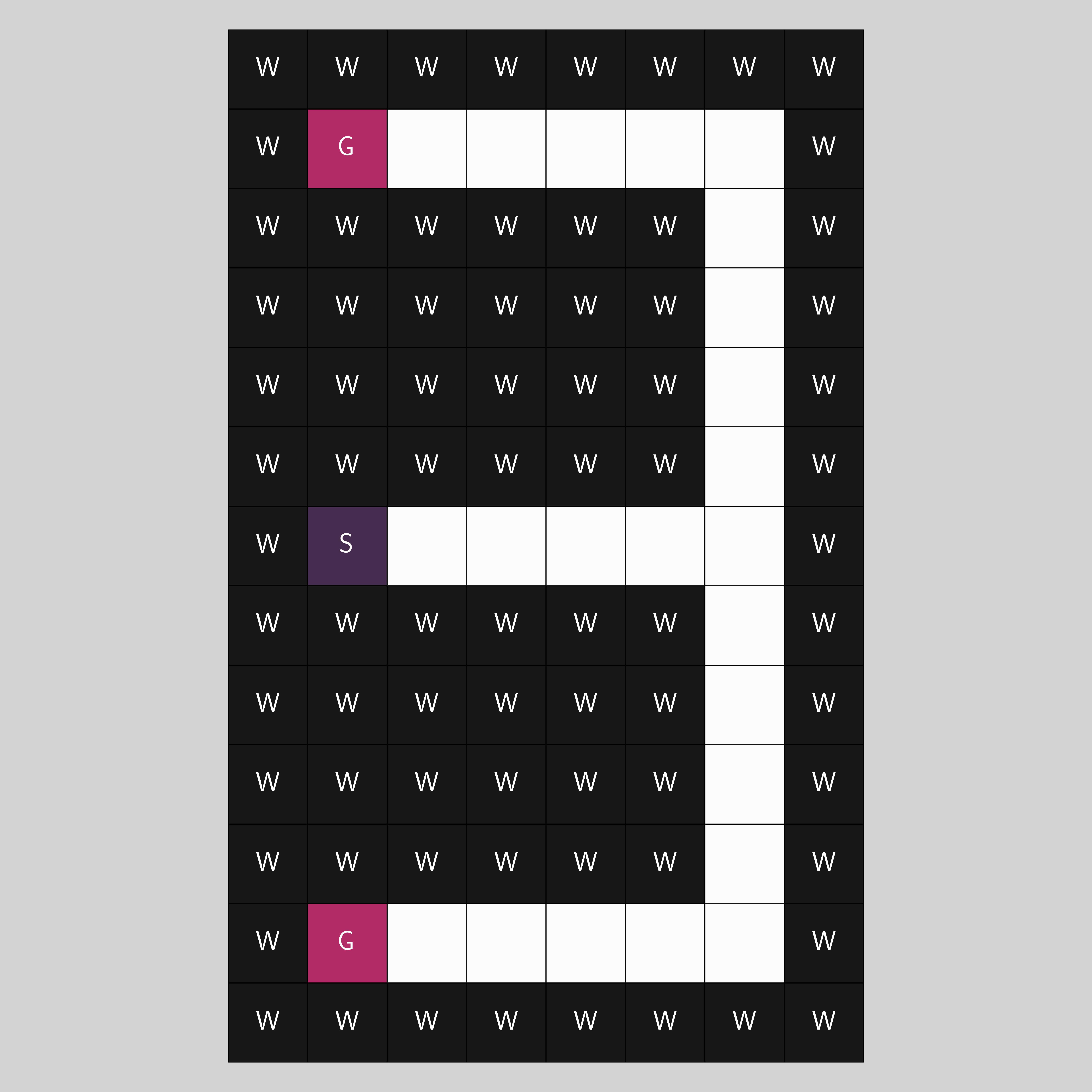}
        \caption{Bidirectional Maze}
        \label{fig:pm:bidirlarge}
    \end{subfigure}
    \hfill
    \begin{subfigure}[t]{0.34\textwidth}
        \includegraphics[height=3.9cm]{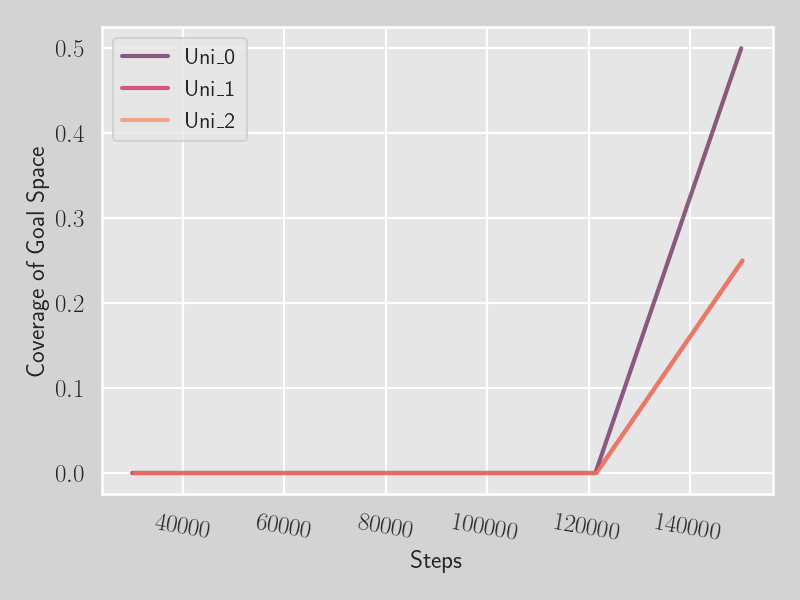}
        \caption{Uni}
        \label{fig:pm:bidir:uni:coverage}
    \end{subfigure}
    \hfill
    \begin{subfigure}[t]{0.34\textwidth}
        \includegraphics[height=3.9cm]{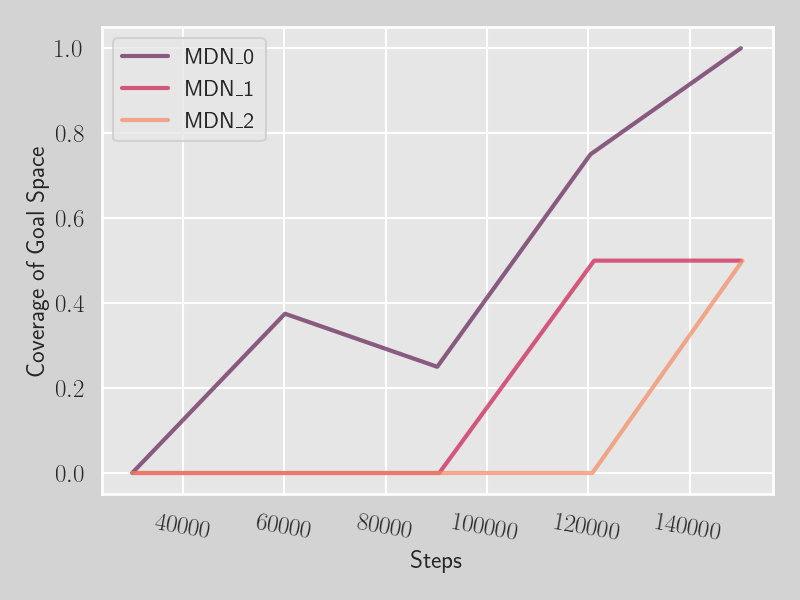}
        \caption{PCL}
        \label{fig:pm:bidir:prc:coverage}    
    \end{subfigure}
    \caption{Bidirectional Maze Coverage}
    \label{fig:pm:bidir:coverage}
\end{figure}

\begin{figure}[htbp]
    \begin{subfigure}[t]{0.245\textwidth}
        \includegraphics[height=3.9cm]{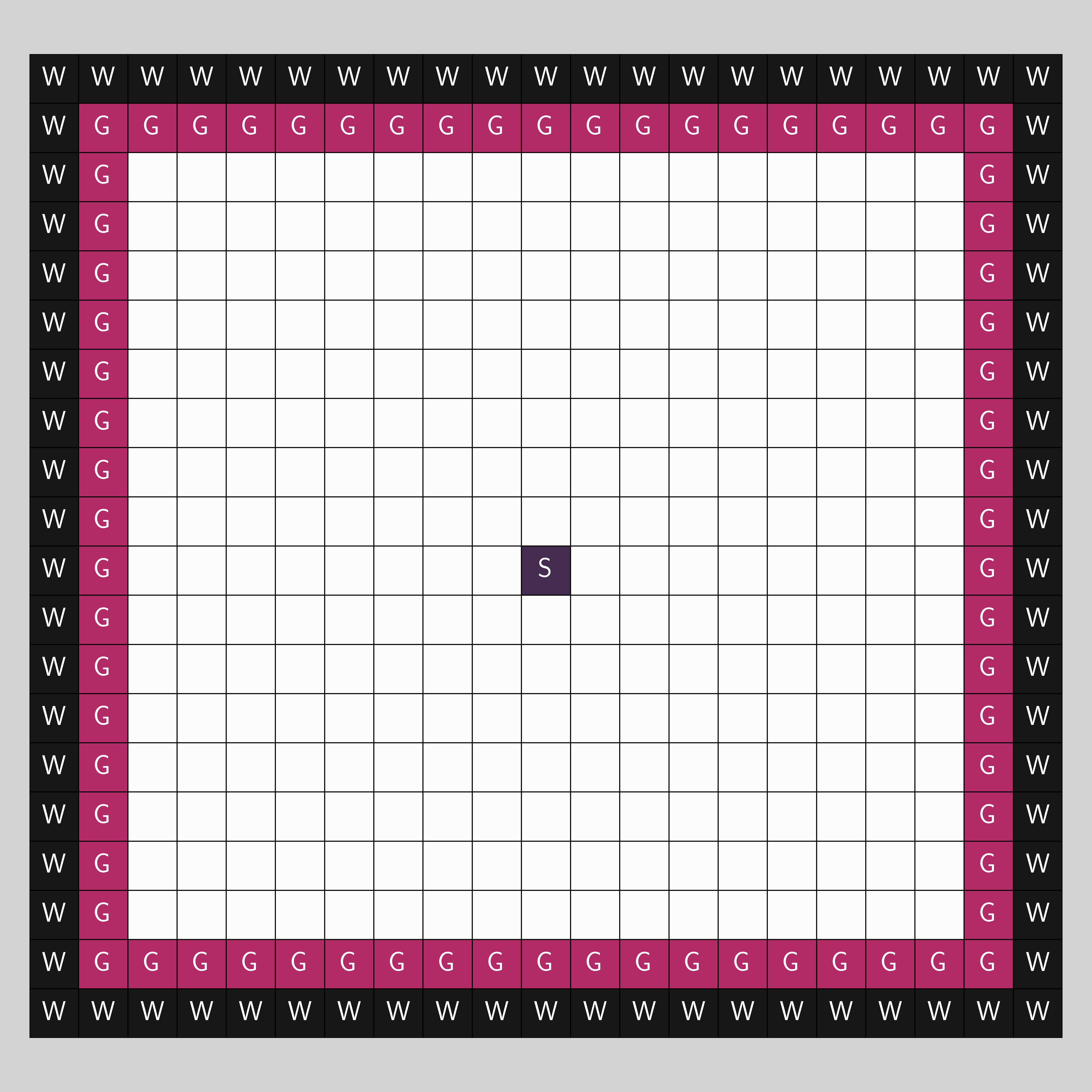}
        \caption{21x21 Square Maze}
        \label{fig:pm:21x21}
    \end{subfigure}
    \hfill
    \begin{subfigure}[t]{0.34\textwidth}
        \includegraphics[height=3.9cm]{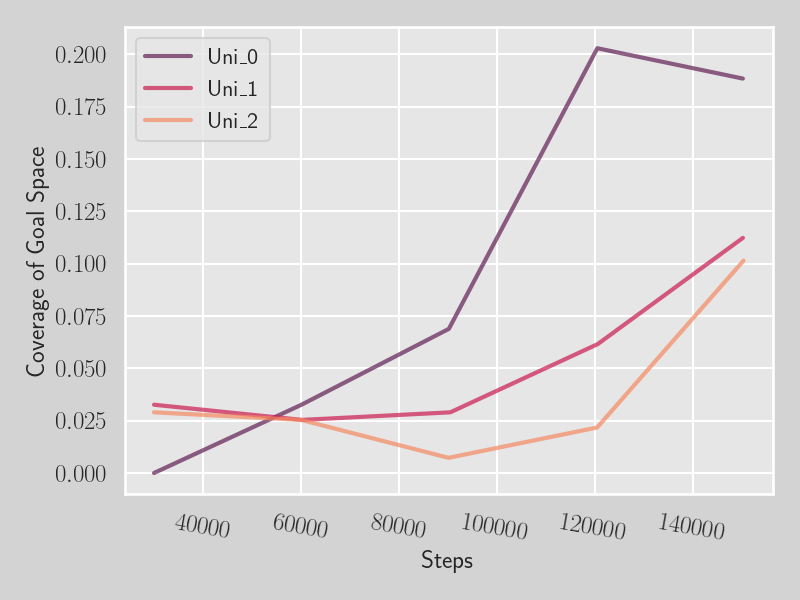}
        \caption{Uni}
        \label{fig:pm:21x21:uni:coverage}
    \end{subfigure}
    \hfill
    \begin{subfigure}[t]{0.34\textwidth}
        \includegraphics[height=3.9cm]{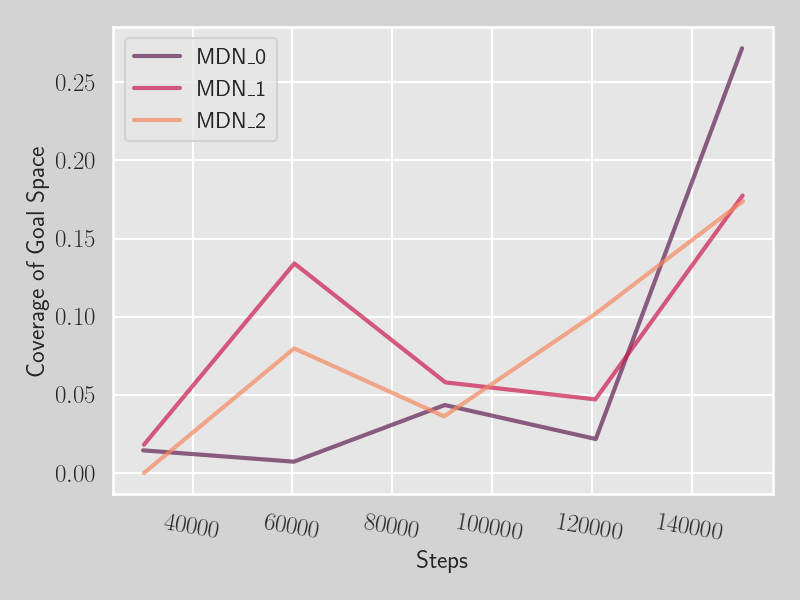}
        \caption{PCL}
        \label{fig:pm:21x21:prc:coverage}    
    \end{subfigure}
    \caption{21x21 Square Maze Coverage}
    \label{fig:pm:21x21:coverage}
\end{figure}

Figures~\ref{fig:pm:bidirlarge} and~\ref{fig:pm:21x21} shows the two mazes that we use to evaluate PCL and the uniform method. S shows the start locations, G shows the goal locations, and W shows the walls. 

The maze in Figure~\ref{fig:pm:bidirlarge} assesses the curriculum's capacity to navigate indirectly to the goal over a long horizon, and the maze in Figure~\ref{fig:pm:21x21} assesses the curriculum's ability to reach many goals. 

Figure~\ref{fig:pm:bidir:coverage} shows the coverage of the PCL and uniform curricula in the bidirectional maze, see Tables~\ref{tab:SMDNUBIDIRLARGE600STEP040325} and~\ref{tab:SUNIBIDIRLARGE600STEP050325} for hyperparameters. PCL is able to achieve a coverage of 1 in one run and 0.5 in the other two. The uniform curriculum is able to achieve 0.5 in one run and 0.25 in the other two. Both curricula were optimised for forty runs, demonstrating that the PCL is able to achieve a higher coverage than the uniform method. This indicates that the PCL is able to learn more efficiently and has increased performance on longer time horizon tasks. 

We can see that the PCL curriculum is able to achieve 0.272 coverage whereas the uniform curriculum achieves 0.188 in the 21x21 square maze as shown in Figure~\ref{fig:pm:21x21:coverage}, see Tables~\ref{tab:21x21SMDN030225} and~\ref{tab:21x21SUNI030225}. There are 72 goals in the 21x21 maze, so this correlates to achieving 79 out of 288 goals for the PCL curriculum and 54 out of 288 goals for the Uniform curriculum. The PCL trends are also relatively similar, further indicating that the PCL is providing appropriate goals for the agent to learn. These plots demonstrate that the PCL assists when learning a diverse set of goals. 

\subsection{Selection Strategies and Adaptive Quantiles}
\begin{figure}[htbp]
    \centering
    \begin{subfigure}[t]{0.29\textwidth}
        \centering
        \includegraphics[height=3.5cm]{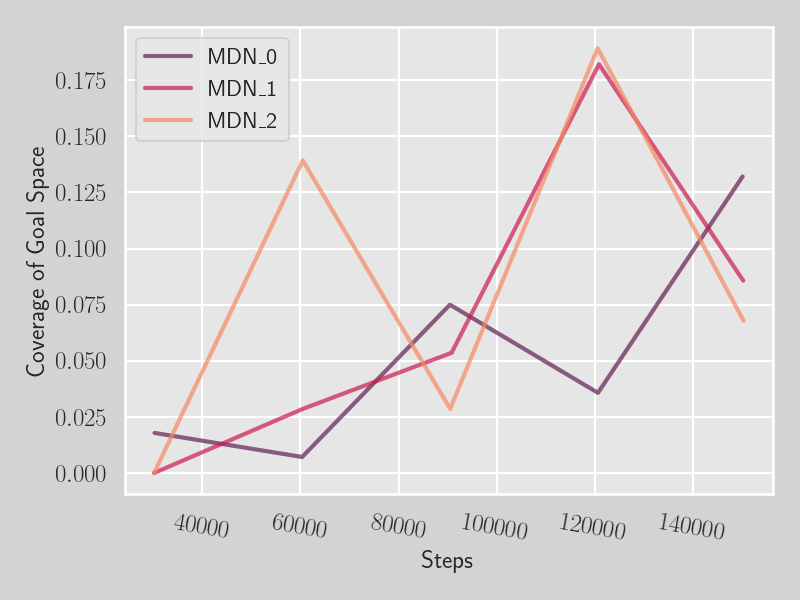}
        \caption{Weighted}
        \label{fig:pm:21x21:w:coverage}
    \end{subfigure}
    \hfill
    \begin{subfigure}[t]{0.29\textwidth}
        \centering
        \includegraphics[height=3.5cm]{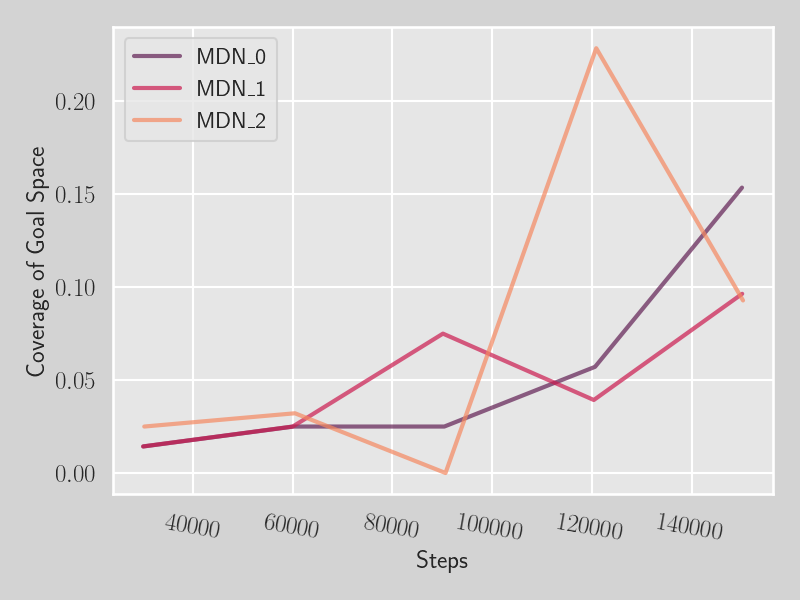}
        \caption{Multiweighted}
        \label{fig:pm:21x21:mw:coverage}    
    \end{subfigure}
    \hfill
    \begin{subfigure}[t]{0.29\textwidth}
        \centering
        \includegraphics[height=3.5cm]{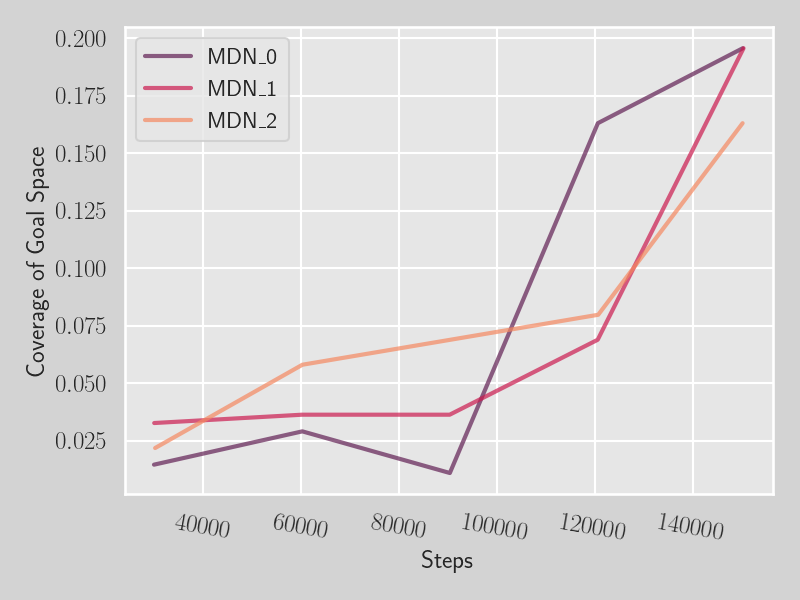}
        \caption{Adaptive Quantiles}
        \label{fig:pm:21x21:a:coverage}
    \end{subfigure}
    \begin{subfigure}[t]{0.29\textwidth}
        \centering
        \includegraphics[height=3.5cm]{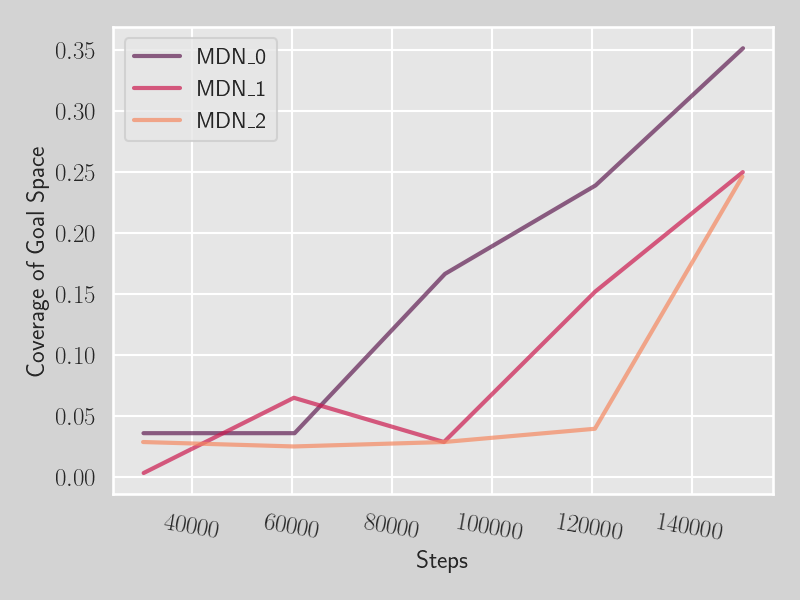}
        \caption{Weighted Adaptive Quantiles}
        \label{fig:pm:21x21:aw:coverage}    
    \end{subfigure}
    \hfill
    \begin{subfigure}[t]{0.29\textwidth}
        \centering
        \includegraphics[height=3.5cm]{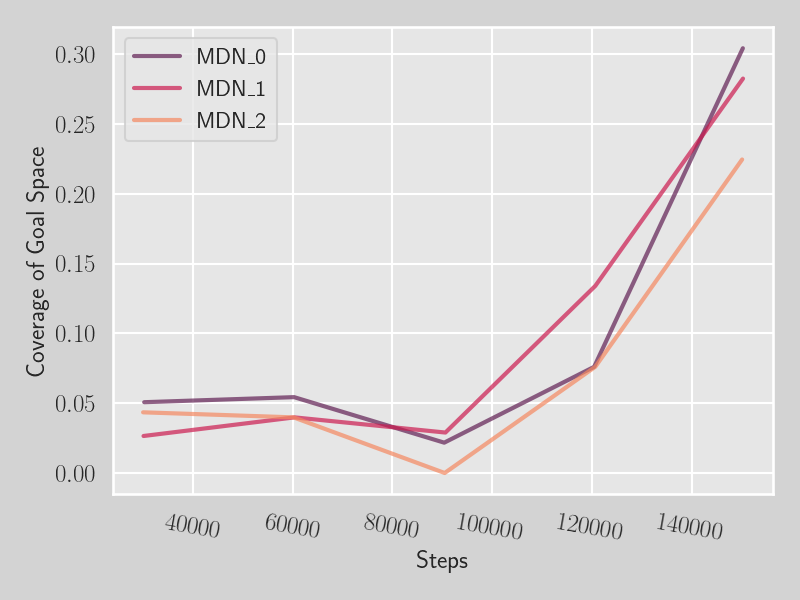}
        \caption{Multiweighted Adaptive Quantiles}
        \label{fig:pm:21x21:amw:coverage}
    \end{subfigure}
    \caption{21x21 Square Maze Selection Strategy and Adaptive Quantiles Coverage}
    \label{fig:pm:21x21:selectionstrategies:coverage}
\end{figure}

In this section we will explore the effect of the selection strategies and adaptive quantiles on the performance of the PCL. We will use the 21x21 maze as it is the most complex and has the most goals. We test the \textbf{weighted}, and \textbf{multiweighted} selection strategies with and without adaptive quantiles, and the \textbf{uniform} strategy with adaptive quantiles. The results are shown in Figure~\ref{fig:pm:21x21:selectionstrategies:coverage}.

When compared to the \textbf{uniform} strategy, the \textbf{weighted} strategy is more exploitative selecting results with higher likelihood of success, the \textbf{multiweighted} strategy combines features to maximise information, the adaptive quantiles encourage exploration when reaching goals and exploitation when not. In Figure~\ref{fig:pm:21x21:selectionstrategies:coverage}, we can see that the \textbf{weighted} and \textbf{multiweighted} selection strategies, and \textbf{adaptive quantiles} achieve coverages of 0.154, 0.200, and 0.196 respectively showing a degradation in performance when compared to the \textbf{uniform} strategy, see Tables~~\ref{tab:21x21MDNW},~\ref{tab:21x21MDNMW}, and~\ref{tab:21x21SMDNADA270225} for hyperparameters. However, when combined the \textbf{weighted} selection strategy with \textbf{adaptive quantiles} achieve a coverage of 0.351, and the \textbf{multiweighted} selection strategy with \textbf{adaptive quantiles} achieves a coverage of 0.304 both outperforming the \textbf{uniform} selection strategy, see Table~\ref{tab:21x21SMDNADAMULTIW0502255} for hyperparameters. The \textbf{weighted} selection strategy with \textbf{adaptive quantiles} suggests that being more exploitative within adaptive quantiles is beneficial when learning many goals with no obstacles, see Table~\ref{tab:21x21SMDNADAW050225} for hyperparameter values.

\section{Conclusion}
We present a novel probabilistic curriculum learning method that utilises SVI and a parametric probabilistic model. The problem is formalised such that the probability density values are used as a surrogate probability metric through quantiles to evaluate the likelihood of reaching a goal given a state and action. 

We show that PCL, Algorithm~\ref{alg:goal:autogoalMDN}, is able to improve learning efficiency, generalise better across multiple goals, and improve performance in longer time horizon tasks when compared to a uniform baseline curriculum. The benefits of PCL are particularly highlighted in the point robot navigation tasks with longer time horizons or many goals. 

Additionally, various selection strategies are explored with and without adaptive quantiles. This demonstrates that there is scope for flexibility within the algorithm depending on the task and environment at hand. Future work will explore the use of other probabilistic models, and the use of other selection strategies.

The advantage of our algorithm is that the probabilistic model is able to both generate and evaluate goals. This allows for a flexible and efficient automatic task generation method. Additionally, we do not constrain out model to narrow distributions or require specific initialisations. We use a deep mixture density network as our probabilistic model, but this could be expanded to other models such as normalising flows. Our custom loss function assists in preventing overfitting and modal collapse which can be problematic where data is incomplete and collected during training like reinforcement learning.

\newpage
\bibliography{neurips}

\newpage
\appendix

\section{Appendix / supplemental material}

\subsection{Hyperparameter Bounds}
\label{app:hyperparams}
\begin{table}[H]
	\centering
	\begin{tabularx}{\textwidth}{|X|X|X|X|}
		\hline
		\textbf{Name} & \textbf{Lower Bound} & \textbf{Upper Bound} & \textbf{Length} \\
		\hline
		Training Frequency (MDN) & 2 & 10 & 1 \\
		\hline
		Number of Mixtures & 6 & 12 & 1 \\
		\hline
		Layers (MDN) & 64 & 1024 & [3, 4] \\
		\hline
		Learning Rate (MDN) & 0.0001 & 1 & 1 \\
		\hline
		$\mathbf{\lambda_1}$ & 0.85 & 2 & 1 \\
		\hline
		$\mathbf{\lambda_2}$ & 0.1 & 0.5 & 1 \\
		\hline
		$\mathbf{\lambda_3}$ & 0.85 & 2 & 1 \\
		\hline
		$\beta_1$ & 0.0 & 2.0 & 1 \\
		\hline
		$\beta_2$ & 0.0 & 2.0 & 1 \\
		\hline
		$\beta_3$ & 0.0 & 2.0 & 1 \\
		\hline
		Number of Samples & 800 & 1200 & 1 \\
		\hline
		$\mathbf{Q_{lower}}$ & 0.01 & 0.6 & 1 \\
		\hline
		$\mathbf{Q_{upper}}$ & 0.61 & 1 & 1 \\
		\hline
		Batch Size (MDN) & 128 & 1024 & 1 \\
		\hline
		Training Frequency (SAC) & 6 & 16 & 1 \\
		\hline
		Batch Size (SAC) & 700 & 1000 & 1 \\
		\hline
		Layers (SAC) & 100 & 800 & [3, 4] \\
		\hline
		Learning Rate (SAC) & 4e-06 & 0.001 & 1 \\
		\hline
	\end{tabularx}
	\caption{Hyperparameter Bounds for Experiments}
	\label{tab:SACWeightedMDNExperiment}
\end{table}

The length indicates the number of numbers generated between the upper and lower bounds. In the case of these experiments, the layers of the neural networks are given a length of 3,4 indicating that there will be 3 to 4 layers in the network. e.g. for the PCL the number of layers is 3 to 4 with each layer containing 64 to 1024 neurons. 

\subsection{Hyperparameter Values For Experiments}
\label{app:hyperparams:values}

\begin{table}[H]
	\centering
	\begin{tabularx}{\textwidth}{|X|X|X|X|}
		\hline
		\textbf{Parameter} & \textbf{exp\_1} & \textbf{exp\_2} & \textbf{exp\_3} \\
		\hline
		$\mathbf{Q_{lower}}$ & 0.216 & 0.118 & 0.33 \\
		\hline
		$\mathbf{Q_{upper}}$ & 0.997 & 0.911 & 0.997 \\
		\hline
		$\mathbf{\lambda_1}$ & 1.49 & 1.65 & 1.5 \\
		\hline
		$\mathbf{\lambda_2}$ & 0.195 & 0.102 & 0.356 \\
		\hline
		$\mathbf{\lambda_3}$ & 1.89 & 1.19 & 1.98 \\
		\hline
		Batch Size (MDN) & 212 & 135 & 246 \\
		\hline
		Batch Size (SAC) & 999 & 914 & 973 \\
		\hline
		Layers (MDN) & [720, 1008, 244, 315] & [1009, 582, 1016, 228] & [507, 945, 332, 106] \\
		\hline
		Layers (SAC) & [114, 694, 469, 312] & [194, 314, 383, 267] & [102, 360, 522, 125] \\
		\hline
		Learning Rate (MDN) & 0.269 & 0.117 & 0.264 \\
		\hline
		Learning Rate (SAC) & 0.000994 & 0.000793 & 0.000995 \\
		\hline
		Number of Mixtures & 12 & 10 & 12 \\
		\hline
		Number of Samples & 970 & 1.09e+03 & 1.06e+03 \\
		\hline
		Training Frequency (MDN) & 2 & 4 & 2 \\
		\hline
		Training Frequency (SAC) & 14 & 15 & 16 \\
		\hline
	\end{tabularx}
	\caption{{Hyperparameters for PCL DC Motor Experiments}}
	\label{tab:SACMDNDCMOTOR270125}
\end{table}
\begin{table}[H]
	\centering
	\begin{tabularx}{\textwidth}{|X|X|X|X|}
		\hline
		\textbf{Parameter} & \textbf{exp\_1} & \textbf{exp\_2} & \textbf{exp\_3} \\
		\hline
		Batch Size (SAC) & 980 & 904 & 987 \\
		\hline
		Layers (SAC) & [388, 590, 628] & [262, 349, 632] & [593, 637, 634] \\
		\hline
		Learning Rate (SAC) & 0.000997 & 0.000984 & 0.000981 \\
		\hline
		Training Frequency (SAC) & 15 & 15 & 16 \\
		\hline
	\end{tabularx}
	\caption{{Hyperparameters for Uniform Curriculum DC Motor Experiments}}
	\label{tab:SACUNIDCMOTOR270125}
\end{table}
\begin{table}[H]
	\centering
	\begin{tabularx}{\textwidth}{|X|X|X|X|}
		\hline
		\textbf{Parameter} & \textbf{exp\_1} & \textbf{exp\_2} & \textbf{exp\_3} \\
		\hline
		$\mathbf{Q_{lower}}$ & 0.408 & 0.256 & 0.498 \\
		\hline
		$\mathbf{Q_{upper}}$ & 0.927 & 0.684 & 0.78 \\
		\hline
		$\mathbf{\lambda_1}$ & 1.89 & 1.24 & 1.62 \\
		\hline
		$\mathbf{\lambda_2}$ & 0.252 & 0.304 & 0.217 \\
		\hline
		$\mathbf{\lambda_3}$ & 0.869 & 1.17 & 0.924 \\
		\hline
		Batch Size (MDN) & 483 & 134 & 887 \\
		\hline
		Batch Size (SAC) & 798 & 907 & 793 \\
		\hline
		Layers (MDN) & [831, 392, 715] & [795, 133, 670] & [500, 960, 740] \\
		\hline
		Layers (SAC) & [678, 209, 574, 148] & [798, 417, 466, 558] & [203, 348, 211] \\
		\hline
		Learning Rate (MDN) & 0.0194 & 0.233 & 0.0455 \\
		\hline
		Learning Rate (SAC) & 0.000405 & 0.000396 & 0.000261 \\
		\hline
		Number of Mixtures & 9 & 10 & 7 \\
		\hline
		Number of Samples & 1.01e+03 & 1.1e+03 & 1.06e+03 \\
		\hline
		Training Frequency (MDN) & 5 & 2 & 9 \\
		\hline
		Training Frequency (SAC) & 7 & 6 & 8 \\
		\hline
	\end{tabularx}
	\caption{{Hyperparameters for PCL Bidirectional Maze Experiments}}
	\label{tab:SMDNUBIDIRLARGE600STEP040325}
\end{table}
\begin{table}[H]
	\centering
	\begin{tabularx}{\textwidth}{|X|X|X|X|}
		\hline
		\textbf{Parameter} & \textbf{exp\_1} & \textbf{exp\_2} & \textbf{exp\_3} \\
		\hline
		Batch Size (SAC) & 813 & 809 & 723 \\
		\hline
		Layers (SAC) & [785, 191, 676, 140] & [543, 172, 595] & [124, 280, 764] \\
		\hline
		Learning Rate (SAC) & 0.00057 & 0.000621 & 0.000798 \\
		\hline
		Training Frequency (SAC) & 16 & 9 & 13 \\
		\hline
	\end{tabularx}
	\caption{{Hyperparameters for Uniform Curriculum Bidirectional Maze Experiments}}
	\label{tab:SUNIBIDIRLARGE600STEP050325}
\end{table}
\begin{table}[H]
	\centering
	\begin{tabularx}{\textwidth}{|X|X|X|X|}
		\hline
		\textbf{Parameter} & \textbf{exp\_1} & \textbf{exp\_2} & \textbf{exp\_3} \\
		\hline
		$\mathbf{Q_{lower}}$ & 0.286 & 0.173 & 0.371 \\
		\hline
		$\mathbf{Q_{upper}}$ & 0.992 & 0.917 & 0.921 \\
		\hline
		$\mathbf{\lambda_1}$ & 1.98 & 1.74 & 1.91 \\
		\hline
		$\mathbf{\lambda_2}$ & 0.413 & 0.311 & 0.308 \\
		\hline
		$\mathbf{\lambda_3}$ & 1.28 & 1.17 & 1.14 \\
		\hline
		Batch Size (MDN) & 495 & 608 & 373 \\
		\hline
		Batch Size (SAC) & 998 & 949 & 946 \\
		\hline
		Layers (MDN) & [69, 742, 1024] & [918, 130, 721] & [903, 761, 722] \\
		\hline
		Layers (SAC) & [211, 306, 762] & [201, 800, 478] & [216, 798, 300] \\
		\hline
		Learning Rate (MDN) & 0.193 & 0.473 & 0.45 \\
		\hline
		Learning Rate (SAC) & 0.000215 & 0.000858 & 0.000963 \\
		\hline
		Number of Mixtures & 6 & 6 & 6 \\
		\hline
		Number of Samples & 827 & 832 & 833 \\
		\hline
		Training Frequency (MDN) & 2 & 9 & 9 \\
		\hline
		Training Frequency (SAC) & 9 & 8 & 8 \\
		\hline
	\end{tabularx}
	\caption{{Hyperparameters for PCL 21x21 Square Maze Experiments}}
	\label{tab:21x21SMDN030225}
\end{table}
\begin{table}[H]
	\centering
	\begin{tabularx}{\textwidth}{|X|X|X|X|}
		\hline
		\textbf{Parameter} & \textbf{exp\_1} & \textbf{exp\_2} & \textbf{exp\_3} \\
		\hline
		Batch Size (SAC) & 889 & 964 & 906 \\
		\hline
		Layers (SAC) & [684, 671, 390] & [480, 715, 459, 645] & [483, 575, 697] \\
		\hline
		Learning Rate (SAC) & 0.000949 & 0.000514 & 0.000938 \\
		\hline
		Training Frequency (SAC) & 6 & 14 & 10 \\
		\hline
	\end{tabularx}
	\caption{{Hyperparameters for Uniform Curriculum 21x21 Square Maze Experiments}}
	\label{tab:21x21SUNI030225}
\end{table}
\begin{table}[H]
	\centering
	\begin{tabularx}{\textwidth}{|X|X|X|X|}
		\hline
		\textbf{Parameter} & \textbf{exp\_1} & \textbf{exp\_2} & \textbf{exp\_3} \\
		\hline
		$\mathbf{Q_{lower}}$ & 0.383 & 0.48 & 0.589 \\
		\hline
		$\mathbf{Q_{upper}}$ & 0.761 & 0.78 & 0.815 \\
		\hline
		$\mathbf{\lambda_1}$ & 1.1 & 1.32 & 1.32 \\
		\hline
		$\mathbf{\lambda_2}$ & 0.332 & 0.244 & 0.277 \\
		\hline
		$\mathbf{\lambda_3}$ & 1.81 & 1.81 & 1.99 \\
		\hline
		Batch Size (MDN) & 632 & 459 & 431 \\
		\hline
		Batch Size (SAC) & 847 & 706 & 808 \\
		\hline
		Layers (MDN) & [503, 885, 264] & [96, 670, 301] & [689, 984, 404] \\
		\hline
		Layers (SAC) & [715, 432, 375] & [518, 246, 344] & [763, 395, 326] \\
		\hline
		Learning Rate (MDN) & 0.361 & 0.89 & 0.993 \\
		\hline
		Learning Rate (SAC) & 0.000958 & 0.000765 & 0.00082 \\
		\hline
		Number of Mixtures & 9 & 6 & 11 \\
		\hline
		Number of Samples & 1.11e+03 & 1.12e+03 & 1.2e+03 \\
		\hline
		Training Frequency (MDN) & 7 & 5 & 4 \\
		\hline
		Training Frequency (SAC) & 13 & 14 & 16 \\
		\hline
	\end{tabularx}
	\caption{{Hyperparameters for PCL + Weighted 21x21 Square Maze Experiments}}
	\label{tab:21x21MDNW}
\end{table}
\begin{table}[H]
	\centering
	\begin{tabularx}{\textwidth}{|X|X|X|X|}
		\hline
		\textbf{Parameter} & \textbf{exp\_1} & \textbf{exp\_2} & \textbf{exp\_3} \\
		\hline
		$\beta_1$ & 1.96 & 1.98 & 1.5 \\
		\hline
		$\beta_2$ & 1.05 & 0.188 & 0.0188 \\
		\hline
		$\beta_3$ & 0.496 & 0.62 & 0.427 \\
		\hline
		$\mathbf{Q_{lower}}$ & 0.457 & 0.469 & 0.216 \\
		\hline
		$\mathbf{Q_{upper}}$ & 0.848 & 0.835 & 0.752 \\
		\hline
		$\mathbf{\lambda_1}$ & 1.73 & 1.78 & 1.65 \\
		\hline
		$\mathbf{\lambda_2}$ & 0.197 & 0.374 & 0.198 \\
		\hline
		$\mathbf{\lambda_3}$ & 1.03 & 0.997 & 1.31 \\
		\hline
		Batch Size (MDN) & 543 & 542 & 172 \\
		\hline
		Batch Size (SAC) & 954 & 937 & 951 \\
		\hline
		Layers (MDN) & [244, 1016, 395] & [230, 133, 444] & [113, 603, 552] \\
		\hline
		Layers (SAC) & [649, 760, 133] & [683, 764, 288] & [379, 625, 795, 423] \\
		\hline
		Learning Rate (MDN) & 0.641 & 0.609 & 0.69 \\
		\hline
		Learning Rate (SAC) & 0.000905 & 0.000843 & 0.000991 \\
		\hline
		Number of Mixtures & 6 & 8 & 6 \\
		\hline
		Number of Samples & 1.16e+03 & 1.2e+03 & 917 \\
		\hline
		Training Frequency (MDN) & 7 & 9 & 2 \\
		\hline
		Training Frequency (SAC) & 6 & 8 & 6 \\
		\hline
	\end{tabularx}
	\caption{{Hyperparameters for PCL + Multiweighted 21x21 Square Maze Experiments}}
	\label{tab:21x21MDNMW}
\end{table}
\begin{table}[H]
	\centering
	\begin{tabularx}{\textwidth}{|X|X|X|X|}
		\hline
		\textbf{Parameter} & \textbf{exp\_1} & \textbf{exp\_2} & \textbf{exp\_3} \\
		\hline
		$\mathbf{\lambda_1}$ & 0.944 & 0.95 & 0.866 \\
		\hline
		$\mathbf{\lambda_2}$ & 0.315 & 0.282 & 0.226 \\
		\hline
		$\mathbf{\lambda_3}$ & 1.91 & 1.8 & 1.97 \\
		\hline
		Batch Size (MDN) & 438 & 265 & 1.01e+03 \\
		\hline
		Batch Size (SAC) & 917 & 959 & 923 \\
		\hline
		Layers (MDN) & [141, 374, 229] & [370, 350, 454] & [275, 401, 337] \\
		\hline
		Layers (SAC) & [661, 699, 316] & [277, 448, 710] & [140, 395, 777] \\
		\hline
		Learning Rate (MDN) & 0.611 & 0.581 & 0.47 \\
		\hline
		Learning Rate (SAC) & 0.000446 & 0.000552 & 0.000859 \\
		\hline
		Number of Mixtures & 11 & 7 & 6 \\
		\hline
		Training Frequency (MDN) & 10 & 10 & 7 \\
		\hline
		Training Frequency (SAC) & 11 & 15 & 16 \\
		\hline
	\end{tabularx}
	\caption{{Hyperparameters for PCL + Adaptive Quantile 21x21 Square Maze Experiments}}
	\label{tab:21x21SMDNADA270225}
\end{table}
\begin{table}[H]
	\centering
	\begin{tabularx}{\textwidth}{|X|X|X|X|}
		\hline
		\textbf{Parameter} & \textbf{exp\_1} & \textbf{exp\_2} & \textbf{exp\_3} \\
		\hline
		$\beta_1$ & 1.67 & 1.08 & 0.935 \\
		\hline
		$\beta_2$ & 0.068 & 1.79 & 1.83 \\
		\hline
		$\beta_3$ & 0.913 & 0.211 & 0.308 \\
		\hline
		$\mathbf{\lambda_1}$ & 1.6 & 1.5 & 1.44 \\
		\hline
		$\mathbf{\lambda_2}$ & 0.161 & 0.135 & 0.106 \\
		\hline
		$\mathbf{\lambda_3}$ & 1.5 & 1.28 & 1.37 \\
		\hline
		Batch Size (MDN) & 130 & 701 & 620 \\
		\hline
		Batch Size (SAC) & 876 & 774 & 761 \\
		\hline
		Layers (MDN) & [1014, 683, 267, 397] & [588, 351, 519] & [544, 555, 922] \\
		\hline
		Layers (SAC) & [412, 712, 509] & [734, 645, 226, 577] & [754, 423, 157, 634] \\
		\hline
		Learning Rate (MDN) & 0.738 & 0.22 & 0.0693 \\
		\hline
		Learning Rate (SAC) & 0.000684 & 0.000798 & 0.000813 \\
		\hline
		Number of Mixtures & 7 & 9 & 9 \\
		\hline
		Training Frequency (MDN) & 5 & 6 & 6 \\
		\hline
		Training Frequency (SAC) & 8 & 12 & 13 \\
		\hline
	\end{tabularx}
	\caption{{Hyperparameters for PCL + Adaptive Quantile + Multiweighted 21x21 Square Maze Experiments}}
	\label{tab:21x21SMDNADAMULTIW0502255}
\end{table}
\begin{table}[H]
	\centering
	\begin{tabularx}{\textwidth}{|X|X|X|X|}
		\hline
		\textbf{Parameter} & \textbf{exp\_1} & \textbf{exp\_2} & \textbf{exp\_3} \\
		\hline
		$\mathbf{\lambda_1}$ & 1.79 & 1.88 & 1.95 \\
		\hline
		$\mathbf{\lambda_2}$ & 0.198 & 0.265 & 0.149 \\
		\hline
		$\mathbf{\lambda_3}$ & 0.999 & 0.937 & 1.18 \\
		\hline
		Batch Size (MDN) & 643 & 726 & 487 \\
		\hline
		Batch Size (SAC) & 793 & 778 & 875 \\
		\hline
		Layers (MDN) & [891, 79, 168, 304] & [1015, 630, 64, 575] & [607, 624, 546, 372] \\
		\hline
		Layers (SAC) & [759, 571, 605, 267] & [721, 564, 533, 226] & [704, 516, 536, 153] \\
		\hline
		Learning Rate (MDN) & 0.514 & 0.428 & 0.448 \\
		\hline
		Learning Rate (SAC) & 0.000788 & 0.000884 & 0.000206 \\
		\hline
		Number of Mixtures & 8 & 12 & 9 \\
		\hline
		Training Frequency (MDN) & 7 & 6 & 6 \\
		\hline
		Training Frequency (SAC) & 6 & 6 & 6 \\
		\hline
	\end{tabularx}
	\caption{{Hyperparameters for PCL + Adaptive Quantile + Weighted 21x21 Square Maze Experiments}}
	\label{tab:21x21SMDNADAW050225}
\end{table}

\subsection{Goal Probability}
\label{app:goal:probability}
We can check the probability that the goal sits within some volume $\mathbf{V}$ which describes
some acceptable region around the goal that an agent has to reach for it to be considered successful:
\begin{equation}
    P(g_c \in \mathbf{V}|s_t,a_t) = \int_\mathbf{V} p(g_c)dg_c
\end{equation}
As $g_c\in\mathbb(R)^N$ this can be expressed as:
\begin{equation}
    \label{eq:goals:independent_probs}
    P(g_{c_1},...,g_{c_N}\in\mathbf{V|s_t,a_t}) = \int_\mathbf{V} p_{g_{c_1},...,g_{c_N}}(g_{c_1},...,g_{c_N}|s_t,a_t)dg_{c_1},...,g_{c_N}
\end{equation}
If we assume that the goal dimensions are independent of each other than we can further simplify this to be:
\begin{equation}
    \begin{split}
        P(g^c_{t}\in \mathbf{V}|s_t,a_t) & = \prod^N_{i=1}P(g_{c_i}\in \mathbf{V}|s_t,a_t) \\
                                   & = \prod^N_{i=1}\int_{\mathbf{V}}p(g_{c_i}|s_t,a_t)dg_{c_i}
    \end{split}
\end{equation} 
For example we may have some hyper-rectangle, $\mathbf{A}$ that is characterised by some matrix 
$\mathbf{E} \in \mathbb{R}^{2\times N}$ where $N$ is the number of goal dimensions:
\begin{equation}
    \label{eq:goal:probs}
    \begin{split}
        P(g^c_{t}\in \mathbf{A}|s_t,a_t) & = \prod^N_{i=1}P(\mathbf{E}_{1,i}\leq g_{c_i}\leq \mathbf{E}_{2,i}|s_t,a_t) \\
                                   & = \prod^N_{i=1}\int_{\mathbf{E}_{1,i}}^{\mathbf{E}_{2,i}}p(g_{c_i}|s_t,a_t)dg_{c_i}
    \end{split}
\end{equation}

\subsection{Gaussian Probability Proof}
\label{app:goal:gaussianproof}
If the model as a Gaussian mixture model then we know: \begin{equation}
    \label{eq:goal:MVN}
    g_c \sim \sum_{j=1}^K\phi_j(s_t,a_t) \mathcal{N}(\mu_j(s_t,a_t),\Sigma_j(s_t,a_t))
  \end{equation}
  For simplicity, we will refer to $\phi_j(s_t,a_t)$, $\mu_j(s_t,a_t)$, and $\Sigma_j(s_t,a_t)$ as $\phi_j$, $\mu_j$, and $\Sigma_j$. 
  
  If we sub the pdf of the learnt multivariate Gaussian mixture model eq~\ref{eq:goal:MVN} into eq \ref{eq:goal:probs}:
  \begin{equation}
        P(g^c_{t}\in \mathbf{V}|s_t,a_t) = \int_{\mathbf{V}}\sum_{j=1}^K\phi_j\frac{\exp(-\frac{1}{2}(g^c_{t}-\mu_j)^T\Sigma_j^{-1}(g^c_{t}-\mu_j))}{\sqrt{(2\pi)^N|\Sigma_j|}}dg^c_{t}     
  \end{equation}
  For ease of integration and computation, we can assume that the volume is a hypercube $\mathbf{C}$ centred about $g_c$ with each dimension offset on either side of $g_c$ by $\epsilon>0$. If we also assume 
  all $\Sigma_j$ are diagonal and the goal dimensions are independent then we can perform the following simplification by utilising eq~\ref{eq:goals:independent_probs}:
  \begin{equation}
    \label{eq:goals:hypercube_diag_MVN}
    P(g^c_{t}\in \mathbf{C}|s_t,a_t) = \sum_{j=1}^K\phi_j\frac{1}{\sqrt{(2\pi)^N|\Sigma_j|}}\prod_{i=1}^N\int_{g_{c_i}-\epsilon}^{g_{c_i}+\epsilon}\exp(-\frac{(g_{c_i}-\mu_{j_i})^2}{2\Sigma_{j_{ii}}})dg_{c_i}
  \end{equation}
  We can then let $u_i=\frac{(g_{c_i}-\mu_{j_i})}{\sqrt{2}\sqrt{\Sigma_{j_{ii}}}}$ which gives $dg_{c_i} = \sqrt{2}\sqrt{\Sigma_{j_{ii}}}du$. The new upper and lower bounds are then given by 
  $u_{i_{ub}}=\frac{g_{c_i}+\epsilon-\mu_{j_i}}{\sqrt{2}\sqrt{\Sigma_{j_{ii}}}}$ and $u_{i_{lb}}=\frac{g_{c_i}-\epsilon-\mu_{j_i}}{\sqrt{2}\sqrt{\Sigma_{j_{ii}}}}$. Substituting these into 
  eq~\ref{eq:goals:hypercube_diag_MVN} yields:
  \begin{equation}
    \label{eq:goal:hypercube_diag_MVN_u}
    P(g^c_{t}\in \mathbf{C}|s_t,a_t) = \sum_{j=1}^K\phi_j\frac{1}{\sqrt{(2\pi)^N|\Sigma_j|}}\prod_{i=1}^N\sqrt{2}\sqrt{\Sigma_{j_{ii}}}\int_{u_{i_{lb}}}^{u_{i_{ub}}}\exp(-u_i^2)du_i
  \end{equation}
  We can then utilise the Gaussian error function:
  \begin{equation}
    \label{eq:goals:erf}
    erf(x) = \frac{2}{\sqrt{\pi}}\int_0^xe^{-t^2} 
  \end{equation}
  Subbing eq~\ref{eq:goals:erf} into eq~\ref{eq:goal:hypercube_diag_MVN_u}:
  \begin{equation}
    \begin{split}
        P(g^c_{t}\in \mathbf{C}|s_t,a_t) &= \sum_{j=1}^K\phi_j\frac{1}{\sqrt{(2\pi)^N|\Sigma_j|}}\prod_{i=1}^N\frac{\sqrt{\pi}\sqrt{\Sigma_{j_{ii}}}}{\sqrt{2}}(erf(u_{i_{ub}})-erf(u_{i_{lb}}))\\
                                   &= \sum_{j=1}^K\phi_j\frac{1}{\sqrt{(2\pi)^N|\Sigma_j|}}\prod_{i=1}^N\frac{\sqrt{\pi}\sqrt{\Sigma_{j_{ii}}}}{\sqrt{2}}(erf(\frac{g_{c_i}+\epsilon-\mu_{j_i}}
                                   {\sqrt{2}\sqrt{\Sigma_{j_{ii}}}})-erf(\frac{g_{c_i}-\epsilon-\mu_{j_i}}{\sqrt{2}\sqrt{\Sigma_{j_{ii}}}}))\\
    \end{split}
  \end{equation}
\end{document}